%% file: main.tex
\documentclass[accepted]{preprint} %

\usepackage[american]{babel}

\usepackage{natbib} %
\usepackage{mathtools} %
\usepackage{booktabs} %
\usepackage{tikz} %

\usepackage{hyperref}
\usepackage{graphicx}
\usepackage{booktabs}       %
\usepackage{amsfonts}
\usepackage{amssymb}
\usepackage{amsthm}
\usepackage{thmtools, thm-restate}

\usepackage{algorithm}
\usepackage[noend]{algpseudocode}
\usepackage{xcolor}

\DeclareMathOperator*{\argmin}{\arg\!\min}
\DeclareMathOperator*{\eqmark}{\stackrel{?}{=}}

\title{Tree Search in DAG Space with Model-based \\Reinforcement Learning for Causal Discovery}

\author[1]{\href{mailto:<v.darvariu@ucl.ac.uk>}{Victor-Alexandru Darvariu}{}}
\author[1]{Stephen Hailes}
\author[1,2]{Mirco Musolesi}
\affil[1]{%
    Department of Computer Science\\
    University College London
}
\affil[2]{%
    Department of Computer Science and Engineering\\
    University of Bologna
}
  
\begin{document}
\maketitle

\begin{abstract}
Identifying causal structure is central to many fields ranging from strategic decision-making to biology and economics. In this work, we propose CD-UCT, a model-based reinforcement learning method for causal discovery based on tree search that builds directed acyclic graphs incrementally. We also formalize and prove the correctness of an efficient algorithm for excluding edges that would introduce cycles, which enables deeper discrete search and sampling in DAG space. The proposed method can be applied broadly to causal Bayesian networks with both discrete and continuous random variables.  We conduct a comprehensive evaluation on synthetic and real-world datasets, showing that CD-UCT substantially outperforms the state-of-the-art model-free reinforcement learning technique and greedy search, constituting a promising advancement for combinatorial methods.

\end{abstract}

\input{tex/1_intro}

\input{tex/2_background}
\input{tex/3_methods}
\input{tex/4_experiments}

\input{tex/5_results}
\input{tex/6_conclusion}

\begin{acknowledgements}
This work was supported by the Turing’s Defence and Security programme through a partnership with the UK government in accordance with the framework agreement between GCHQ \& The Alan Turing Institute. Experiments for this paper were carried out on the UCL Department of Computer Science High Performance Computing (HPC) cluster.
\end{acknowledgements}

\bibliography{bibliography}

\newpage

\onecolumn

\input{tex/supplementary}

\end{document}

%% file: tex/1_intro.tex
\section{INTRODUCTION}

Causal graphs are probabilistic graphical models representing causal dependencies between random variables. They are widely used in many empirical sciences and practical scenarios~\citep{friedman2000using,morgan2015counterfactuals} and can enable estimating treatment effects, performing interventions on variables, and answering counterfactual queries~\citep{pearl2009causality}. Determining causal structure from observational data is a fundamental task that arises, for instance, when performing randomized controlled trials is impossible or unethical.

In particular, score-based methods seek to find one of the optimal causal graphs with respect to a score function~\citep{peters2017elements} such as the Bayesian Information Criterion~\citep{schwarz1978estimating}. Usually, the problem is framed as determining a Directed Acyclic Graph (DAG), which eliminates feedback loops. This has been shown to be NP-hard~\citep{chickering2004large}, driving decades of research into heuristic or exact methods.
Reinforcement Learning (RL) has recently been proposed as a means of navigating this search space, motivated by advances on other NP-hard problems. A notable method is RL-BIC~\citep{zhu2020causal} -- a model-free, continuous approach for one-shot DAG generation. However, its performance on real datasets and scalability on large causal graphs remain limited. 

In this paper, we propose Causal Discovery Upper Confidence Bound for Trees (CD-UCT), a practical yet rigorous RL method for causal discovery based on \textit{incremental causal graph construction}. Unlike RL-BIC, it is model-based and performs tree search in a discrete space using a granular Markov Decision Process (MDP) formulation. The core intuition is that access to this model substantially improves the quality of exploration over model-free methods. The advantages of model-based RL have been demonstrated for solving a variety of other problems including the construction of undirected graphs~\citep{darvariu2023planning}. 
The proposed method enjoys the flexibility of RL and is applicable across many types of Bayesian networks, data generation models, and score functions. Furthermore, it is theoretically grounded: given it is based on UCT, it converges to the optimal action as the number of samples grows to infinity~\citep{kocsis_bandit_2006}, akin to the properties of Greedy Equivalence Search~\citep{chickering2002optimal}.

The valid action space of the RL agent must exclude edges that would introduce cycles. As the depth and breadth of the tree grow, explicitly checking for cycles becomes prohibitively expensive. To address this key challenge, we propose an incremental algorithm for efficiently tracking cycle-inducing possible edges as the construction progresses. We prove its correctness and empirically show that it results in a speedup of more than an order of magnitude on the largest graphs tested compared to na\"{\i}ve cycle checks. 

\definecolor{myred}{RGB}{215, 77, 46}
\definecolor{myblue}{RGB}{126, 166, 224}

\begin{figure*}[t]
   \centering
  \includegraphics[width=0.94\textwidth,trim={0 0 0 0},clip=true]{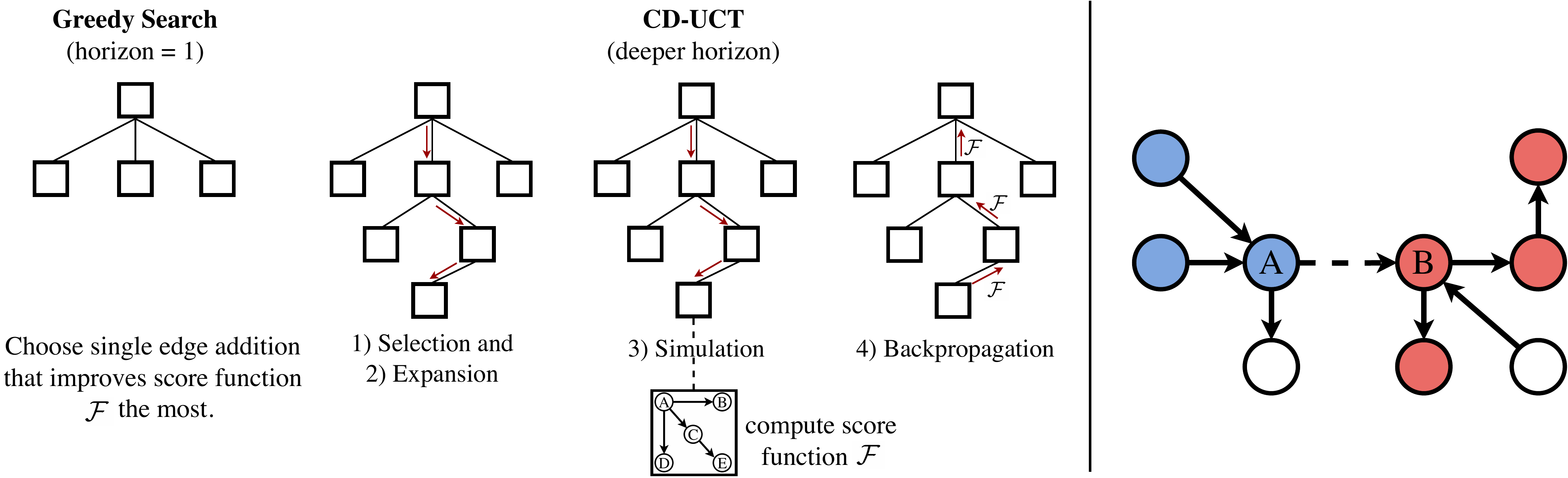}
     \caption{Left: schematic comparison of Greedy Search and CD-UCT, which build shallow and deeper trees respectively to search in DAG space. Right: we propose an incremental algorithm to exclude cycle-inducing edges. It relies on the insight that, after adding edge $A \to B$, connecting a \textcolor{myred}{\textbf{descendant}} of $B$ to an \textcolor{myblue}{\textbf{ancestor}} of $A$ would introduce a cycle in all subsequent timesteps. An illustration of all the algorithm steps can be found in the Supplementary Material.} 
  \label{cdrl_illustration}
\end{figure*}

We evaluate CD-UCT on several real-world and synthetic benchmarks, showing consistently better performance than the state-of-the-art solution in RL for causal discovery (namely, RL-BIC) and Greedy Search in all settings tested. The flexibility of our method is demonstrated by its compatibility with Bayesian networks with both discrete and random variables using a variety of score functions. We also study the impacts of the simulation budget, graph density, and dataset size on algorithm performance. Finally, we analyze the scalability of CD-UCT, which is shown to substantially exceed that of RL-BIC by scaling to graphs with up to $d=50$ nodes.

%% file: tex/2_background.tex
\section{RELATED WORK}

\textbf{Combinatorial methods.} The goal of combinatorial methods is to find the discrete causal structure that optimizes a given score function. Identifying the optimal DAG (we note that there may be more than one) is known to be NP-hard~\citep{chickering2004large}. As such, exact methods based, for example, on dynamic programming~\citep{koivisto2004exact} or LP formulations~\citep{jaakkola2010learning} are fairly limited in their ability to scale. Alternatively, the problem can be solved approximately, with some such methods scaling successfully to Bayesian networks with thousands~\citep{scanagatta2015learning} or even millions~\citep{ramsey2017million} of variables. Local search methods such as Greedy Search in DAG space~\citep{chickering95a} or Markov Equivalence Classes~\citep{meek1997graphical,chickering2002optimal} have been widely successful. In particular, the Greedy Equivalence Search (GES) method, which belongs to the latter category, is proven to find the global optimum in the limit of infinite data.\footnote{Interestingly, while Christopher Meek's PhD thesis~\citep{meek1997graphical}, which first proposed GES, acknowledges that Greedy Search is somewhat simplistic and other types of search algorithms should be considered,  to the best of our knowledge, the bulk of work has concentrated on GES-like methods.}
This guarantee does not hold in the finite data regime, however. The key pathology that Greedy Search can suffer from is the ``horizon effect''~\citep{berliner1973some}, arising due to its shallow search tree. It may perform myopic choices as it is unable to explore longer trajectories that contain sets of edges with better score when considered jointly. We illustrate this in Figure 1 and include a detailed discussion in Section~\ref{subsec:greedy} of the Supplementary Material.

\textbf{Continuous methods.} A recent line of work employs continuous optimization, relying on a smooth characterization of acyclicity proposed by~\cite{zheng2018dags}, which is used as a constraint in a continuous optimization program. Other works build further in this direction by modeling non-linear relationships with neural networks~\citep{yu2019dag,lachapelle2020gradient}. Such techniques, however, are not guaranteed to return DAGs due to the non-convexity of the objective function~\citep{ng2022convergence}. %

\textbf{Order-based methods.} Another important category of approaches adopts a two-step procedure. Firstly, a causal ordering is determined based on the data. Secondly, the method seeks the best-scoring graph that is consistent with the ordering. This can lead to a substantially smaller and regular search space~\citep{friedman2003being}; however, errors made in the first step may have a downstream impact on the second. There are several notable approaches based on this idea~\citep{friedman2003being,buehlmann2014cam}, including several recent works based on continuous optimization~\citep{charpentier2022differentiable,zantedeschi2023dag}.

\textbf{RL for combinatorial optimization.} The present paper belongs to the emerging area of reinforcement learning for combinatorial optimization~\citep{bello_neural_2016,khalil_learning_2017}. Particularly relevant are those works that consider the construction of graphs as an incremental decision-making problem~\citep{dai_adversarial_2018,you_graph_2018,li_learning_2018,darvariu2021goal}. Our approach is most closely related to \cite{darvariu2023planning}, which proposes a Monte Carlo Tree Search method for constructing spatial networks.

\textbf{RL-BIC.} Introduced by~\cite{zhu2020causal}, RL-BIC is a score-based continuous method that relies on the acyclicity characterization proposed by~\citet{zheng2018dags} in conjunction with RL. Since the score function is used to provide rewards, its differentiability is not required. It features an encoder-decoder architecture trained using the policy gradient for one-shot DAG generation. The best-scoring graph found during training is returned as output.
 

%% file: tex/3_methods.tex
\section{METHODS}

\subsection{Problem formulation}

Let $G=(V,E)$ denote a DAG with $d$ nodes and $m$ edges. Each node $v_i \in V$ corresponds to a Random Variable (RV) $x_i$ that may be discrete or continuous, while edges $e_{i,j}$ indicate directional causal relationships. Let $\mathbf{x}$ denote the vector $(x_1, x_2, \dots x_d)$ of RVs distributed according to $p(\mathbf{x})$. Let $\mathbf{Pa}(x_i)$ denote the parent set of $x_i$ in the causal graph, i.e., RVs $x_k$ s.t. $e_{k,i} \ \in E$. Variables $x_i$ are assumed to be independent of other RVs given their parent set: 

\begin{align}
p\left(x_1, \ldots, x_d\right)=\prod_{i=1}^d p\left(x_i \mid \mathbf{Pa}(x_i) \right).
\end{align} 

This suffices to represent the problem with discrete RVs. For continuous RVs, we follow the general Structural Equation Model (SEM) of~\cite{pearl2009causality}, in which RVs are generated according to $x_i = f_i(\mathbf{Pa}(x_i))$ where $f_i$ represents some (typically unknown) function. Many subclasses of SEM exist, with varying assumptions and properties. For example, in the Additive Noise Model~\citep{hoyer2008nonlinear}, variables are generated by applying nonlinear functions to parents and adding arbitrary and jointly independent noise.

\textbf{Causal discovery.} Given a dataset $\mathbf{X} \in \mathbb{R}^{n \times d}$ of $n$ $d$-dimensional observations, the goal is to identify the true underlying DAG $G$. In \textit{score-based} methods, this is formulated as the optimization of a score function $\mathcal{F}$. Letting $\mathbb{D}^{(d)}$ denote the set of DAGs with $d$ nodes, the problem can be formalized as finding one of the graphs $G^*$ satisfying
\begin{align}
G^*&=\argmin_{G \in \mathbb{D}^{(d)}} {\mathcal{F}(G)},
\end{align}

\textbf{Score functions.} We consider the Bayesian Information Criterion (BIC), which is known to be consistent and decomposable (noting, however, that the proposed method does not rely on the latter property). For discrete RVs, the BIC score function $\mathcal{F}_{\mathrm{DV}}$ can be computed as~\citep{koller2009probabilistic}:

\begin{align}
\mathcal{F}_{\mathrm{DV}}(G) &= -\left(\ell \ (G|\mathbf{X}) - \frac{1}{2} m \log n\right), 
\end{align}

where $\ell(G|\mathbf{X})$ denotes the log-likelihood of the model parameters (i.e., the graph structure) w.r.t. the observed data. For continuous RVs, following~\cite{zhu2020causal}, we consider two instantiations of BIC based on residuals and treat variances as heterogeneous ($\mathcal{F}_{\mathrm{HV}}$) and equal ($\mathcal{F}_{\mathrm{EV}}$):

\begin{align}
\mathcal{F}_{\mathrm{HV}}(G) &= \sum_{i=1}^d\left(n \log \left(\mathrm{RSS}_i / n\right)\right)+m \log n \\
\mathcal{F}_{\mathrm{EV}}(G) &= n d \log \left(\left(\sum_{i=1}^d \operatorname{RSS}_i\right) /(n d)\right)+ m \log n,
\end{align}

where $\mathrm{RSS}_i$ is the residual sum of squares obtained by regressing $x_i$ on its parent variables. Unless otherwise stated, we use Gaussian Process regression~\citep{williams2006gaussian} for fitting the data and computing the residuals. Intuitively, the search problem can be thought of as finding the DAG that best explains the observed data, with a penalty on the number of edges encouraging sparsity.

\subsection{DAG construction as MDP}~\label{subsec:mdp}

We next describe our formulation of the problem of finding an optimal DAG as an MDP. Unlike RL-BIC, whose graph generation is one-shot, we frame the problem incrementally, i.e., as a decision-making process. We add edges one by one to a (possibly empty) initial graph until an edge budget $b$ is exhausted. To maintain a manageable $\mathcal{O}(d)$ action space, we decompose the addition of an edge into two separate decision-making steps. The MDP components are defined as follows:

\textbf{State.} A state $S_t$ is formed of a tuple $\left(G_t,\{ \sigma_t \}\right)$ containing the DAG $G_t=(V, E_t)$ and a singleton containing an \textit{edge stub} $\sigma_t$. At even timesteps ($t\ \text{mod}\ 2 = 0$), $\sigma_t$ is equal to the empty set $\varnothing$. At odd timesteps, $\sigma_t$ is equal to $v_k$, where $v_k \in V$ is the node that was selected in the previous timestep, from which a directed edge must be built.

\textbf{Action.} An action $A_t$ corresponds to the \textit{selection of a node} in $V$. To ensure that the acyclicity property is maintained, actions that would introduce cycles must be excluded from the action space. Let $\textsc{IsCyclic}(G)$ be a function that, given a DAG $G$, outputs a Boolean value corresponding to whether the graph contains at least one cycle. We now define the set of edges $\mathcal{C}_t$, whose introduction after time $t$ induces a cycle. Furthermore, let $\mathcal{K}_t(v_i)$ denote those \textit{connectable} nodes $v_j$ such that the candidate edge $e_{i,j}$ is valid:

\begin{align}\label{eq:cycsetdef}
\mathcal{C}_t = \{e_{i,j} \notin E_t \ &|\  \textsc{IsCyclic} (V, E_t \cup \{ e_{i, j}\}) \}, \\ 
\mathcal{K}_t(v_i) = \{v_j\ &|\  \ e_{i,j} \notin \mathcal{C}_t \}.
\end{align}

The set of available actions containing the nodes that may be selected is defined as below, where $\text{deg}^+(v_i)$ denotes the out-degree of node $v_i$. The first clause states that maximally connected nodes or those from which only cycle-inducing edges originate cannot be selected as the edge stub. The second clause, in which $v_k$ denotes the node selected at the previous timestep, forbids actions that would lead to the construction of a cycle-inducing or already-existing edge.
\begin{align}\label{eq:actionspace}
\mathcal{A}(S_t)=
  \begin{cases}
    \{ v_{k} \in V &|\ \text{deg}^+(v_{k}) < d-1 \land |\mathcal{K}_t(v_k)| > 0\}, \\ \ &\text{if}\  t\ \text{mod}\ 2 = 0 \\
    \{ v_{l} \in V &|\ e_{k, l} \notin E_t \land v_l \in \mathcal{K}_t(v_k) \}, \\ \  &\text{otherwise.}
  \end{cases}
\end{align}

\textbf{Transitions.} The dynamics is deterministic, meaning that, from a state $s$, there is a single state $s'$ that can be reached with probability $1$. When transitioning from an even timestep, the model ``marks'' the node corresponding to the selected action as the edge stub so that it forms part of the next state. Otherwise, it adds the selected edge to the topology of the next state, and resets the edge stub.

\textbf{Reward.} The reward $R_t$ is defined as the negative of the score so as to reconcile the reward maximization paradigm of RL with minimizing the score function. Namely,
  \begin{align}
  R_t =
  \begin{cases}
    - \mathcal{F}(G_t),\  \text{if}\   t=2b \\
    0,\  \text{otherwise.}
  \end{cases}
\end{align}

\subsection{Incremental algorithm for detecting cycle-inducing edges}

Computing the set $\mathcal{C}_t$ (Equation~\ref{eq:cycsetdef}) is required for determining the valid action space (Equation~\ref{eq:actionspace}). One strategy is to explicitly evaluate the membership condition for all edges that do not exist in the current edge set (which we refer to as \textit{candidate} edges in the remainder of this work). Cycle existence can be determined by creating a copy of the graph with the candidate edge added and running a cycle detection algorithm (e.g., a traversal such as depth-first search).

However, performing the necessary cycle checks for a set of candidate edges explicitly scales as $\mathcal{O}(d^3)$, assuming a sparse graph and DFS traversal. Given this must be performed after every edge addition (i.e., every $2$ MDP timesteps), this is substantially inefficient, especially if sampling longer trajectories for estimating future reward, as it is typically done in tree search. With this approach, deeper searches are not feasible beyond very small graphs. Instead, we propose a means of keeping track of the cycle-inducing candidate edges by leveraging the fact that our MDP formulation is incremental (i.e., it builds the graph edge-by-edge). 

Let us first introduce some key concepts prior to stating our result. For ease of exposition, we denote a coarser-grained timestep $\tau$, which corresponds to $2$ MDP timesteps and advances with each edge addition. We also let $\mathbf{De}(v_i)$ denote the set of \textit{descendants} of $v_i$ (i.e., nodes that can be reached starting from $v_i$ via a directed path);  $\mathbf{An}(v_i)$ denote the set of \textit{ancestors} of $v_i$ (nodes that can reach $v_i$ via a directed path). Both sets are taken to be closed (i.e., they include $v_i$). 

\begin{restatable}{theorem}{cyc}
\label{th:cyc}
Let $G_\tau$ denote a directed acyclic graph and known cycle-inducing candidate edges $\mathcal{C}_\tau$. Given that edge $e_{i,j}$ is chosen for addition at timestep $\tau$ $(e_{i,j} \in E_{\tau+1})$, the set $\mathcal{C}_{\tau+1}$ is equal to $\mathcal{C}_\tau \cup \Phi_{i,j}$, where $\Phi_{i,j} = \{e_{x,y}\notin E_{\tau+1} \ |\ (v_x, v_y) \ \in \mathbf{De}(v_j) \times \mathbf{An}(v_i) \}$.
\end{restatable}

The proof is deferred to Section~\ref{sec:thproof} of the Supplementary Material. The incremental algorithm making use of this update rule is presented in Algorithm~\ref{alg:cyc}, and a full run over multiple timesteps is illustrated in the Supplementary Material. 

A remaining point to address is the initial set $\mathcal{C}_0$. If construction begins from scratch, it trivially holds that the only invalid choice is edge $e_{j,i}$. If starting from an existing graph, $\mathcal{C}_0$ can be computed using traversals or, alternatively, applying the update rule in Theorem~\ref{th:cyc} with an arbitrary ordering of the initial edges. This only needs to be performed \textit{once} as an initialization step for the search process. We emphasize that, beyond this, \textit{no graph traversals to determine cycles are needed at any point during execution}. 

\begin{algorithm}[tb]
   \caption{Determining cycle-inducing candidates.}
   \label{alg:cyc}
\begin{algorithmic}[0]
   \State {\bfseries Input:} timestep $\tau$, DAG $G_\tau=(V,E_\tau)$, \\
   prior cycle-inducing candidates $\mathcal{C}_\tau$, \\
   chosen next edge $e_{i,j}$ to add at time $\tau$.
   \State {\bfseries Output:} next cycle-inducing candidates $\mathcal{C}_{\tau+1}$.
   \vspace{0.5\baselineskip}

   \State  \textbf{if} {$\tau=0$} \textbf{then} \Comment{initialize cycle-inducing candidates}
           \State \hskip1.5em \textbf{if} {$|E_\tau|=0$} \textbf{then} $\mathcal{C}_\tau = \{e_{j,i}\}$ 
           \State \hskip1.5em \textbf{else}
           		   \State \hskip3em $\mathcal{C}_\tau = \{ \}$
           		   \State \hskip3em \textbf{for} {$e_{x,y} \notin E_\tau$}
           		   			\State \hskip4.5em \textbf{if} $\textsc{IsCyclic} (V, E_{\tau} \cup \{ e_{x, y}\})$ \textbf{then}
           		   				\State \hskip6em $\textsc{Add}(\mathcal{C}_\tau,e_{x,y})$

   \State $\Phi_{i,j} = \{e_{x,y} \notin E_{\tau+1} \ |\ (v_x, v_y) \ \in \mathbf{De}(v_j) \times \mathbf{An}(v_i) \}$
   \State \textbf{return} $\mathcal{C}_\tau \cup \Phi_{i,j}$

\end{algorithmic}
\end{algorithm}

\subsection{The CD-UCT Method}

The method, which we term Causal Discovery UCT (CD-UCT), is given in pseudocode in Algorithm~\ref{alg:cd-uct}. It builds on the Upper Confidence Bound for Trees (UCT)~\citep{kocsis_bandit_2006} variant of Monte Carlo Tree Search (MCTS). A detailed overview of MCTS is given in Section~\ref{subsec:mcts} of the Supplementary Material, including definitions of the components used in the pseudocode description. 

At a high level, the algorithm proceeds in a loop (lines 9-24) to decide a sequence of actions that define a DAG. Each step of the nested loop (14-20) decides an individual action and proceeds until a budget of simulations is exhausted. Within each step, the algorithm navigates the search tree using its tree policy that balances exploration and exploitation and adds a new node (\textit{Selection} and \textit{Expansion}, line 15); samples valid edges using its simulation policy until a terminal state is reached and the reward can be computed (\textit{Simulation}, 16); and backpropagates the reward to all nodes along the trajectory (\textit{Backpropagation}, 17). The child of the root node with the highest average reward is chosen as the next root (line 21) and the process repeats. Regarding the differences to standard UCT, we highlight the use of the proposed Algorithm~\ref{alg:cyc} to compute the valid action space (10, 11). Furthermore, we note the adoption of the memoization of the best trajectory found during the search (8, 18-20, 22), as in~\citep{zhu2020causal,darvariu2023planning}.  

\begin{algorithm}[tb]
   \caption{Causal Discovery UCT (CD-UCT).}
   \label{alg:cd-uct}
\begin{algorithmic}[1]
   \State {\bfseries Input:} DAG $G_0=(V,E_0)$, score function $\mathcal{F}$, \\
   edge budget $b$, simulation multiplier $b_\text{sims}$, \\
   search horizon $h$.
   \State {\bfseries Output:} actions $A_0, \dots A_{T-1}$ defining \\
   best-scoring DAG $G_T=(V,E_T)$ w.r.t. $\mathcal{F}$.
   \vspace{0.5\baselineskip}

   \State $t=0$, $S_0=(G_0, \{\varnothing\})$, $r_\text{max}= -\infty$ 
   \State compute $\mathcal{C}_t$ using \textbf{Algorithm~\ref{alg:cyc}}
   \State $\text{bestAs}=\textsc{Array}()$, $\text{pastAs} = \textsc{Array}()$

   \State \textbf{loop}
   	  \State \hskip1.5em update $\mathcal{C}_t$ using \textbf{Algorithm~\ref{alg:cyc}}
   	  \State \hskip1.5em compute $\mathcal{A}(S_t)$ using \textbf{Equation~\ref{eq:actionspace}}
      \State \hskip1.5em \textbf{if} $t = 2b\ \textbf{or}\ |\mathcal{A}(S_t)|=0$ \textbf{then} \textbf{return} $\text{bestAs}$
      \State  \hskip1.5em create root node $n_\text{root}$ from $S_t$
      
      \State \hskip1.5em  \textbf{for} {$i=0$ to $(b_{\text{sims}} * d)$}

         \State  \hskip3.0em $n_\text{border}, \text{treeAs} =$ \textsc{TreePolicy}($n_\text{root}$)
         \State  \hskip3.0em $r, \text{outAs} =$ \textsc{SimPolicy}($n_\text{border}, h$) 
         \State  \hskip3.0em \textsc{Backup}($n_\text{border},r$)
         \State  \hskip3.0em \textbf{if} {$r > r_\text{max}$} \textbf{then} 
            \State \hskip4.5em $\text{bestAs} = [\text{pastAs}, \text{treeAs}, \text{outAs}]$
            \State \hskip4.5em $r_\text{max} = r$

      \State \hskip1.5em  $n_\text{child}=$ \textsc{MaxChild}($n_\text{root}$) 
      \State \hskip1.5em  $\textsc{Append}(\text{pastAs}, \textsc{GetAction}(n_\text{child}))$
      \State \hskip1.5em  $t+=1$, $S_t=\textsc{GetState}(n_\text{child})$
   \State \textbf{return} $\text{bestAs}$
\end{algorithmic}
\end{algorithm}

%% file: tex/4_experiments.tex
\section{EXPERIMENTS}~\label{sec:experiments}

\textbf{Datasets with continuous variables.} We evaluate the methods on two real-world tasks in the biological domain: \textit{Sachs}~\citep{sachs2005causal} and \textit{SynTReN}~\citep{van2006syntren}. The former involves determining causal influences of protein and phospholipid components in the signaling pathways of cells ($d=11, m=17, n=853$). The latter concerns the determination of the structure of gene regulatory networks from gene expression data, and consists of 10 transcriptional networks generated by a simulator ($d=20, m\in [20,\ldots, 25],n=500$). Both are widely used in the ML causal discovery literature.

Our focus on realistic data is due to recent work showing that common synthetic benchmarks may be trivial to solve~\citep{reisach2021beware}. This is based on the observation that marginal variance in simulated DAGs with the Additive Noise Model tends to increase along the causal order. Hence, a simple baseline that determines a causal order by sorting on the variances and performs a sparse regression on the predecessors can be competitive with state-of-the-art algorithms on such synthetic graphs. 

Nevertheless, a benefit of synthetic data is that it allows fine-grained control of the generation parameters. With this caveat in mind, we also perform $2$ experiments on graphs with uniformly sampled edges (i.e., Erd{\H{o}}s-R{\'e}nyi). In the first experiment, we consider graphs with $d=10$ and vary the number of edges $m \in \{15,20,25,30,35,40,45\}$ and datapoints $n \in \{10^1, \ldots, 5 * 10^3\}$ generated using GP regression. For the second experiment, we consider a graph with $d=50$, $m=113$ (yielded by an edge probability of $0.1$), and $n=1000$ generated with quadratic regression, which is quicker for graphs of this scale.

\textbf{Datasets with discrete variables.} We also evaluate applicable methods on classic benchmarks in the Bayesian networks literature: \textit{Asia}~\citep{lauritzen1988local} ($d=8, m=8$), \textit{Child}~\citep{spiegelhalter1992learning} ($d=20, m=25$), and \textit{Insurance}~\citep{binder1997adaptive} ($d=27, m=52$). We use $n=1000$ samples for each.

\textbf{Baselines.} We compare the proposed method with RL-BIC~\citep{zhu2020causal}, Greedy Search, Random Search, and Random Sampling -- all of which are directly comparable given our problem formulation. Additionally, for continuous RVs, we examine the performance of the method against the non-combinatorial methods CAM~\citep{buehlmann2014cam}, LiNGAM~\citep{shimizu2006linear}, and NOTEARS~\citep{zheng2018dags}. For discrete RVs, we compare with the exact method GOBNILP~\citep{cussens2012bayesian}. Further details about the baselines are provided in the Supplementary Material.

\begin{table*}[t]
\caption{Results obtained by the methods in the construction phase (top) and after undergoing pruning (bottom). CD-UCT performs best out of all comparable methods for construction, both in terms of rewards and downstream metrics. Post-pruning, CD-UCT performs best on \textit{Sachs} and is competitive with the other methods on \textit{SynTReN}.}
\label{tab:joint-results}
\begin{center}
\begin{small}
\resizebox{1\textwidth}{!}{
\input{results/diffvar_joint_final}
}
\end{small}
\end{center}
\end{table*}

\begin{figure*}[t]
\centering
\includegraphics[width=\textwidth]{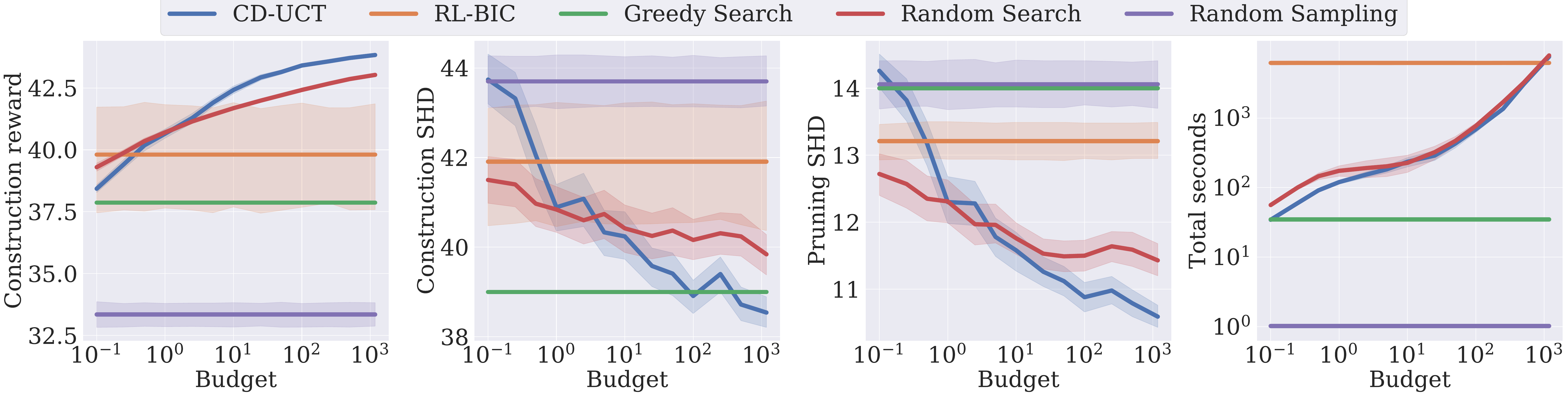} 
\caption{Varying the $b_\text{sims}$ simulation budget parameter on \textit{Sachs}. The subfigures show the construction reward, Structural Hamming Distance (SHD) for construction and pruning, and wall clock time. CD-UCT and Random Search both outperform RL-BIC, even when given $100\times$ fewer score function evaluations.}
\label{fig:budget}
\end{figure*}

\textbf{Construct-then-prune.} We use a two-step procedure for determining the causal graph in the continuous variables case. In the first stage, which we call the \textit{construct} phase, plausible causal relationships are generated. In the second stage, which we call the \textit{pruning} phase, connections are pruned according to some criterion. This paradigm is used by many causal discovery methods~\citep{chickering2002optimal,buehlmann2014cam,zhu2020causal}. Following RL-BIC, we use the procedure proposed in CAM by~\cite{buehlmann2014cam} for the pruning phase, which applies a sparse regression and removes edges corresponding to non-significant relationships. For fairness of comparison, all methods undergo CAM pruning. We note that the results of the \textit{construct} phase are still meaningful as they allow us to compare the effectiveness of the search space exploration conducted by the combinatorial methods.

\textbf{Metrics.} We report the Reward metric, which directly corresponds to the score and is used by the agents directly as the maximization objective. Additionally, we report the True Positive Rate (TPR) -- the fraction of all causal relationships that are correctly identified; False Discovery Rate (FDR) -- the fraction of causal relationships that are predicted by the model that are incorrect; and Structural Hamming Distance (SHD) -- the minimum number of edge additions, deletions, and reversals that are required to transform the output graph into the ground truth causal graph.

\textbf{Experimental procedure and implementation.} We afford CD-UCT and RL-BIC the same number of $4$ hyperparameter configurations tuned on a held-out set of seeds to maximize reward, and use the default hyperparameters for the other methods. We use $100$ runs for the \textit{Sachs} dataset, $20$ each for the \textit{SynTReN} and $d=10$ synthetic graphs, and $50$ for the $d=50$ synthetic graph and the discrete RV graphs. Where applicable, we report means and 95\% confidence intervals\footnote{In a future version, our implementation, data, and instructions will be publicly released in order to ensure full reproducibility of all the reported results. Further details about the implementation are provided in the Supplementary Material.}.

\textbf{Determining budgets.} To ensure fair comparisons between CD-UCT and RL-BIC, we need to align the edge budgets $b$ and score function evaluation budgets $b_{\text{sims}}$. To achieve this, we measure the average number of edges output by RL-BIC in the \textit{construct} phase across multiple runs, yielding $b=49$ edges for \textit{Sachs} and $b=97$ for \textit{SynTReN}. Additionally, we set $b_{\text{sims}}$ to match the $1.28 * 10^6$ evaluations of the score function performed by RL-BIC with the parameters suggested by the authors. We use $b_{\text{sims}}=1178$ on \textit{Sachs} and $b_{\text{sims}}=337$ and $b_{\text{sims}}=33$ for CD-UCT and Random Search respectively on \textit{SynTReN}. The $10\times$ smaller simulation budget for Random Search on the latter dataset is due to its exploration inefficiency leading to longer runtimes (see the ``Runtime analysis'' paragraph in the Supplementary Material). For the synthetic and discrete RV graphs, we set $b$ equal to the true number of edges and $b_{\text{sims}}=1000$.

%% file: results/diffvar_joint_final.tex
\begin{tabular}{ccc|rrrrrrrr}
\toprule
     Phase & Dataset & Metric  &        CD-UCT &         RL-BIC &  Greedy Search &  Random Search & Random Sampling &      CAM &   LiNGAM &  NOTEARS \\
\midrule
 \textbf{Construct} &    \textit{Sachs} &  Reward $\uparrow$ &  43.834\tiny{$\pm0.044$} &   39.807\tiny{$\pm2.209$} &         37.867 &   43.030\tiny{$\pm0.051$} &    33.343\tiny{$\pm0.486$} & --- & --- & --- \\
  &     &     TPR $\uparrow$ &   0.664\tiny{$\pm0.013$} &    0.548\tiny{$\pm0.032$} &          0.588 &    0.620\tiny{$\pm0.022$} &     0.421\tiny{$\pm0.030$} & --- & --- & --- \\
  &     &     FDR $\downarrow$ &   0.770\tiny{$\pm0.005$} &    0.811\tiny{$\pm0.008$} &          0.796 &    0.785\tiny{$\pm0.008$} &     0.854\tiny{$\pm0.010$} & --- & --- & --- \\
  &     &     SHD $\downarrow$ &  38.680\tiny{$\pm0.342$} &   41.910\tiny{$\pm1.383$} &         39.000 &   39.840\tiny{$\pm0.478$} &    43.700\tiny{$\pm0.587$} & --- & --- & --- \\
\midrule
  &  \textit{SynTReN} &  Reward $\uparrow$ &  97.847\tiny{$\pm0.339$} &   73.874\tiny{$\pm4.348$} &         97.700 &   89.181\tiny{$\pm0.311$} &    74.717\tiny{$\pm0.676$} & --- & --- & --- \\
  &   &     TPR $\uparrow$ &   0.376\tiny{$\pm0.021$} &    0.308\tiny{$\pm0.033$} &          0.292 &    0.285\tiny{$\pm0.016$} &     0.256\tiny{$\pm0.014$} & --- & --- & --- \\
  &   &     FDR $\downarrow$ &   0.907\tiny{$\pm0.006$} &    0.932\tiny{$\pm0.005$} &          0.925 &    0.930\tiny{$\pm0.004$} &     0.938\tiny{$\pm0.004$} & --- & --- & --- \\
\rule{0pt}{2ex}  &   &     SHD $\downarrow$ &  96.035\tiny{$\pm1.181$} &  107.387\tiny{$\pm6.930$} &         98.000 &  100.884\tiny{$\pm0.877$} &   102.155\tiny{$\pm0.899$} & --- & --- & --- \\
\midrule[0.175em]
\rule{0pt}{2ex}         \textbf{Prune} &    \textit{Sachs} &     TPR $\uparrow$ &   0.432\tiny{$\pm0.011$} &    0.309\tiny{$\pm0.019$} &          0.235 &    0.373\tiny{$\pm0.014$} &     0.211\tiny{$\pm0.021$} &    0.353 &    0.235 &    0.118 \\
      &     &     FDR $\downarrow$ &   0.258\tiny{$\pm0.018$} &    0.447\tiny{$\pm0.028$} &          0.556 &    0.319\tiny{$\pm0.025$} &     0.577\tiny{$\pm0.039$} &    0.400 &    0.556 &    0.500 \\
      &     &     SHD $\downarrow$ &  10.600\tiny{$\pm0.183$} &   13.210\tiny{$\pm0.284$} &         14.000 &   11.430\tiny{$\pm0.251$} &    14.060\tiny{$\pm0.352$} &   12.000 &   14.000 &   15.000 \\
\midrule
      &  \textit{SynTReN} &     TPR $\uparrow$ &   0.305\tiny{$\pm0.018$} &    0.149\tiny{$\pm0.014$} &          0.264 &    0.231\tiny{$\pm0.014$} &     0.189\tiny{$\pm0.012$} &    0.329 &    0.303 &   $\times$ \\
      &   &     FDR $\downarrow$ &   0.812\tiny{$\pm0.013$} &    0.887\tiny{$\pm0.010$} &          0.839 &    0.865\tiny{$\pm0.009$} &     0.879\tiny{$\pm0.008$} &    0.794 &    0.809 &    $\times$ \\
      &   &     SHD $\downarrow$ &  43.890\tiny{$\pm1.158$} &   43.683\tiny{$\pm1.152$} &         46.100 &   49.352\tiny{$\pm0.990$} &    47.715\tiny{$\pm0.946$} &   41.300 &   41.600 &   $\times$ \\
\bottomrule
\end{tabular}

%% file: tex/5_results.tex
\section{RESULTS}\label{sec:results}

In this section, we present results with the $\mathcal{F}_{\mathrm{DV}}$ score function for discrete variables, and the $\mathcal{F}_{\mathrm{HV}}$ score function for continuous RVs. For completeness, we repeat the main experiments with the $\mathcal{F}_{\mathrm{EV}}$ score function with the more restrictive ``equal variances'' assumption. These results are shown in the Supplementary Material and display similar characteristics. 

\textbf{Main results with continuous variables.} The key results on the \textit{Sachs} and \textit{SynTReN} datasets are shown in Table~\ref{tab:joint-results}. In the construction phase, our method obtains the best results out of the considered combinatorial methods in terms of the rewards received and the metrics. The performance improvements with respect to RL-BIC are substantial. Furthermore, RL-BIC achieves worse results than a Random Search that is afforded a similar number of score function evaluations. 

After pruning, CD-UCT maintains the best results out of all combinatorial methods. It performs the best overall on \textit{Sachs} yielding an SHD of $10.6$, but is outperformed on average by the order-based CAM method on the \textit{SynTReN} graphs. Note that the SHD of RL-BIC on \textit{SynTReN} is lower due to the pruning eliminating more non-significant edges rather than its identification of true relationships, as reflected in the much poorer TPR and FDR. Furthermore, the $\times$ in the NOTEARS column is due to the method not returning DAGs for some of the instances in this dataset, as it may become stuck in local optima~\citep{ng2022convergence}. Wall clock runtimes are shown in the Supplementary Material and demonstrate CD-UCT is over an order of magnitude faster than RL-BIC on \textit{SynTReN} due to its targeted exploration. 

\textbf{Budget analysis.} We conduct an experiment where we vary the simulation budget multiplier afforded to the CD-UCT and Random Search agents for the \textit{Sachs} dataset. These results are shown in Figure~\ref{fig:budget}, in which the largest value on the $x$-axis corresponds to equal simulations to RL-BIC. Beyond a tiny simulation budget, CD-UCT is indeed able to navigate the search space better and find higher-scoring graphs than Random Search. Strikingly, even with a modest simulation multiplier corresponding to $\approx100\times$ less simulations than RL-BIC, both CD-UCT and Random Search yield better reward and SHD. In Section~\ref{subsec:rlbic}, the interested reader can find an additional discussion of the performance of CD-UCT compared to RL-BIC (and, more generally, of model-based approaches compared to model-free ones).

\begin{figure}[t]
\centering
\includegraphics[width=0.65\columnwidth]{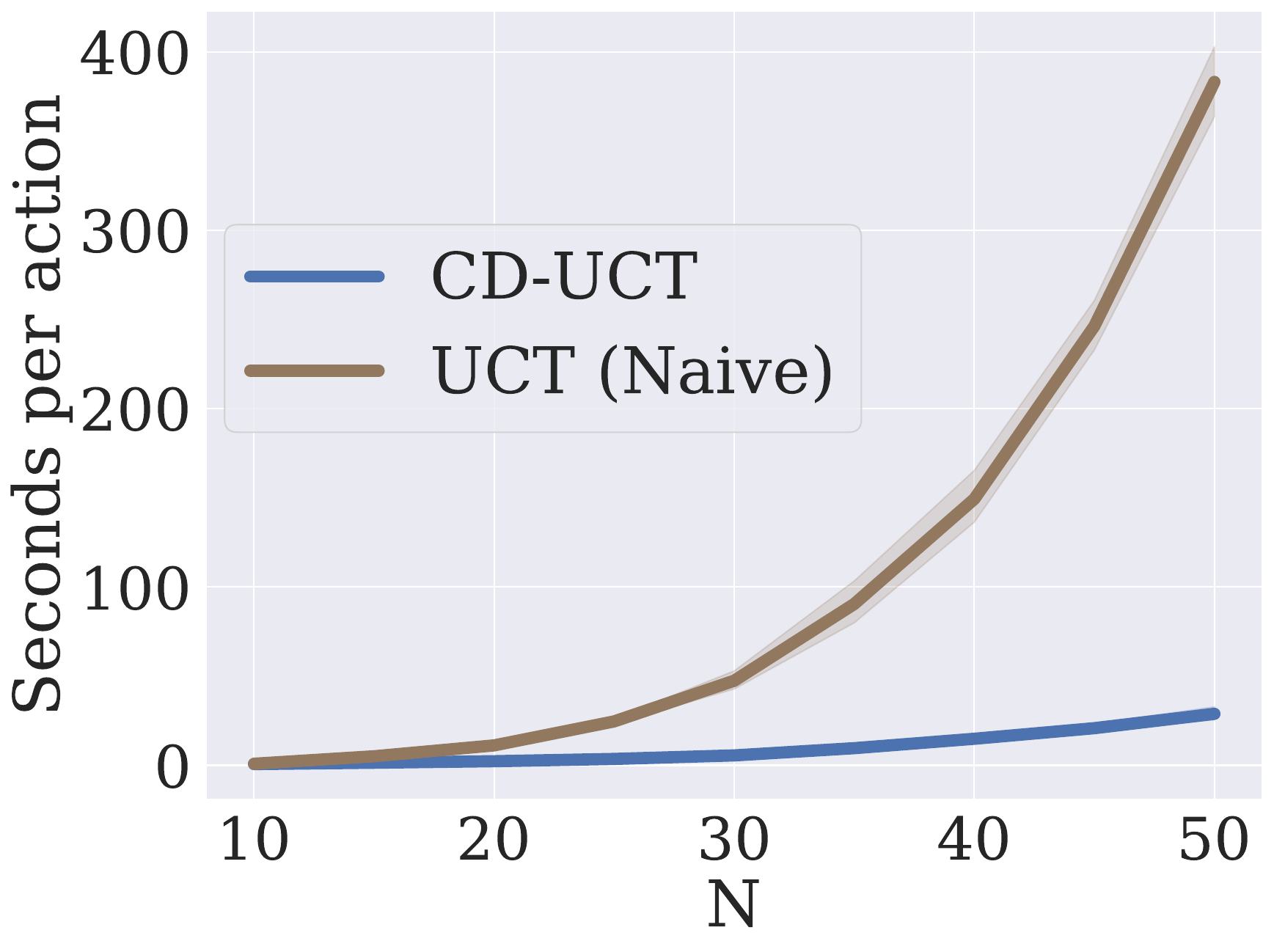} 
\caption{Runtimes of CD-UCT with the incremental Algorithm~\ref{alg:cyc} against a na\"{\i}ve implementation that performs traversals to detect cycles.}%
\label{fig:timings}
\end{figure}

\textbf{Impact of incremental algorithm.} We examine the impact of Algorithm~\ref{alg:cyc} for eliminating cycle-inducing candidates by comparing CD-UCT with a na\"{\i}ve version that performs traversals, all other aspects being equal. This is carried out on graphs with $d \in \{10,20,30,40,50\}$, edge probability of $0.1$, quadratic regression, and $b_{\text{sims}}=10$. We showcase the results of this analysis in Figure~\ref{fig:timings}. The speedup scales super-linearly and improves from a factor of approximately $1.25\times$ for graphs with $d=10$ to $13.25\times$ for graphs with $d=50$. We conclude that this algorithm is a key component for tree search in DAG space with larger graphs.

\begin{table*}[t]
\caption{Results with discrete RVs (top) and $d=50$ synthetic continuous RVs (bottom).}
\label{tab:discrete-results}
\begin{center}
\begin{small}
\resizebox{0.65\textwidth}{!}{
\input{results/diffvar_construct_final_discrete}
}

\vspace{0.3cm}

\resizebox{0.8\textwidth}{!}{
\input{results/diffvar_construct_final_scaleup}
}
\end{small}
\end{center}
\end{table*}

\textbf{Main results with discrete variables.} As shown in Table~\ref{tab:discrete-results}, CD-UCT generally outperforms the other approximate combinatorial methods. The performance difference between CD-UCT and Greedy Search is smaller relative to the continuous RV case. The relationship between Random Search and Greedy Search is also reversed: the latter obtains substantially better results for the larger two graphs. This suggests that this setting favors shorter search horizons. The exclusion of RL-BIC from this experiment is due to its incompatibility with discrete RVs. 

\begin{figure}[t]
\centering
\includegraphics[width=\columnwidth]{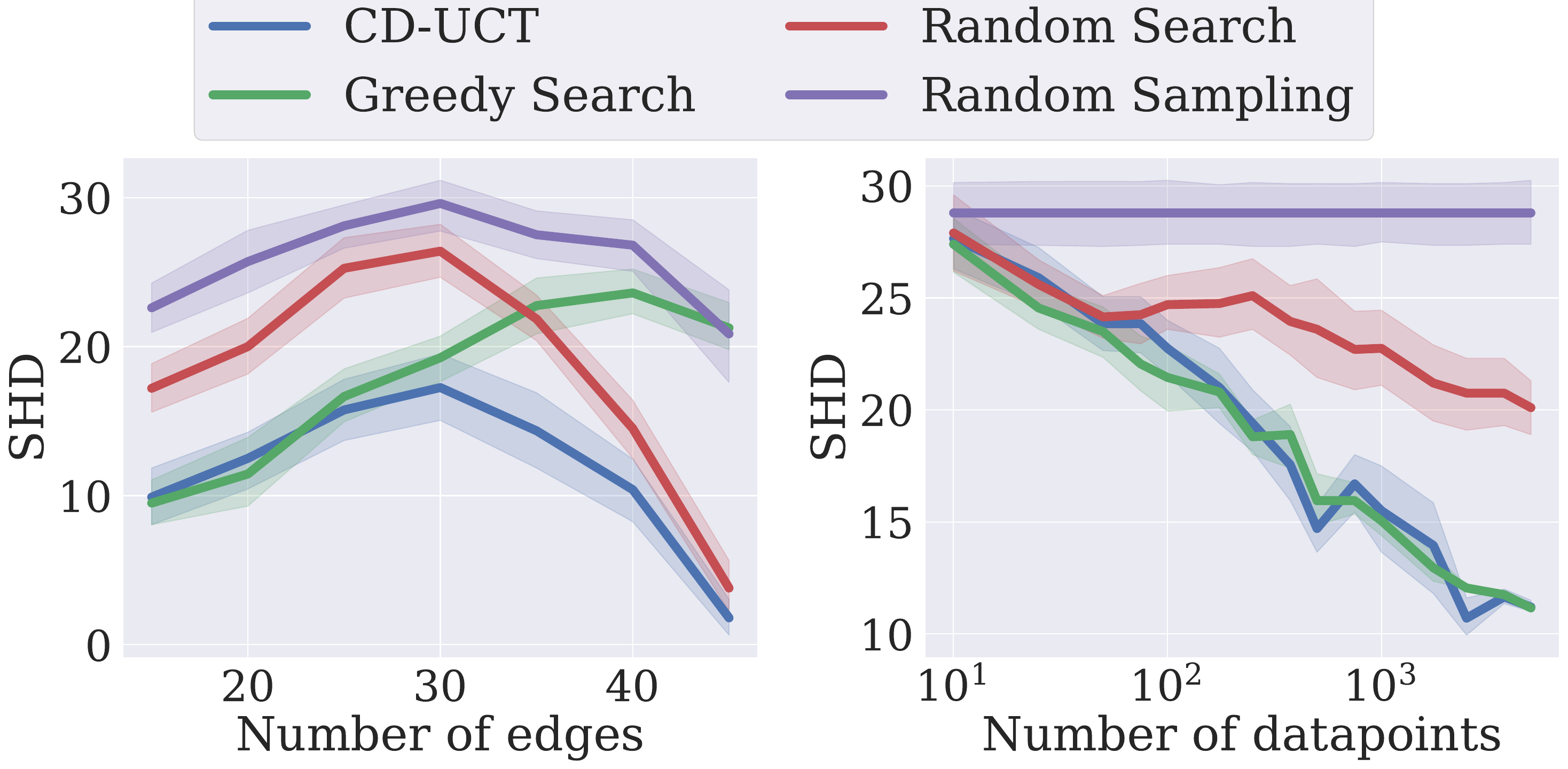} 
\caption{Results with problem instances of varying graph density and number of datapoints.}
\label{fig:synth}
\end{figure}

\textbf{Synthetic graph experiments.} In Figure~\ref{fig:synth}, we examine the impact of the true graph density and dataset size on performance. When the edge count is low (and thus most nodes have one parent), CD-UCT and Greedy Search perform similarly. As the density grows, and determining the relationships edge-by-edge may no longer be accurate, the two curves diverge: CD-UCT improves substantially (as does Random Search), while Greedy Search is akin to randomly sampling a trajectory. Regarding dataset size, performance improves for all methods as the number of datapoints increases. The relative performance of CD-UCT and Greedy Search remains similar, while Random Search plateaus.

\textbf{Scaling to larger graphs.} Our final experiment is performed with a synthetic graph with $d=50$ as previously described. These results are displayed in Table~\ref{tab:discrete-results}. CD-UCT shows statistically significant better performance in terms of reward and metrics with respect to Greedy Search, albeit only marginally for the latter. RL-BIC cannot function with graphs of this size as it needs to model $d^2$ edge probabilities, which is unsurprising given the remarkable increase in wall clock time observed from $d=11$ to $d=20$ as shown in the Supplementary Material. We attribute the better performance of score-based methods relative to CAM, LiNGAM, and NOTEARS to the use of quadratic regression both in the ground truth generation and the scoring function, acting as a strong model class prior.

%% file: results/diffvar_construct_final_discrete.tex
\begin{tabular}{cr|rrrrr}
\toprule
 Dataset  & Metric  &         CD-UCT &  Greedy Search & Random Search & Random Sampling &  GOBNILP \\
\midrule
     \textit{Asia} & Reward $\uparrow$ &  -2.171\tiny{$\pm0.002$} &         -2.177 &  -2.207\tiny{$\pm0.003$} &    -2.727\tiny{$\pm0.044$} &   -2.165 \\
     &    TPR $\uparrow$ &   0.505\tiny{$\pm0.062$} &          0.250 &   0.362\tiny{$\pm0.052$} &     0.128\tiny{$\pm0.033$} &    0.750 \\
      &    FDR $\downarrow$ &   0.495\tiny{$\pm0.062$} &          0.750 &   0.637\tiny{$\pm0.052$} &     0.873\tiny{$\pm0.033$} &    0.250 \\
      &    SHD $\downarrow$ &   5.120\tiny{$\pm0.503$} &          7.000 &   7.620\tiny{$\pm0.520$} &    12.620\tiny{$\pm0.571$} &    3.000 \\
\midrule
    \textit{Child} & Reward $\uparrow$ & -13.036\tiny{$\pm0.026$} &        -13.018 & -15.154\tiny{$\pm0.041$} &   -17.646\tiny{$\pm0.306$} &  -12.972 \\
    &    TPR $\uparrow$ &   0.670\tiny{$\pm0.029$} &          0.640 &   0.174\tiny{$\pm0.018$} &     0.064\tiny{$\pm0.016$} &    0.800 \\
     &    FDR $\downarrow$ &   0.233\tiny{$\pm0.037$} &          0.360 &   0.826\tiny{$\pm0.018$} &     0.936\tiny{$\pm0.016$} &    0.091 \\
     &    SHD $\downarrow$ &   9.580\tiny{$\pm1.133$} &         12.000 &  37.700\tiny{$\pm0.699$} &    45.300\tiny{$\pm0.827$} &    5.000 \\
\midrule
\textit{Insurance} & Reward $\uparrow$ & -14.740\tiny{$\pm0.029$} &        -14.746 & -19.115\tiny{$\pm0.068$} &   -27.093\tiny{$\pm2.220$} &  -14.358 \\
 &    TPR $\uparrow$ &   0.331\tiny{$\pm0.014$} &          0.308 &   0.126\tiny{$\pm0.012$} &     0.078\tiny{$\pm0.010$} &    0.481 \\
 &    FDR $\downarrow$ &   0.596\tiny{$\pm0.038$} &          0.692 &   0.874\tiny{$\pm0.012$} &     0.922\tiny{$\pm0.010$} &    0.265 \\
 &    SHD $\downarrow$ &  54.460\tiny{$\pm2.646$} &         57.000 &  85.080\tiny{$\pm1.167$} &    92.080\tiny{$\pm1.080$} &   28.000 \\
\bottomrule
\end{tabular}

%% file: results/diffvar_construct_final_scaleup.tex
\begin{tabular}{c|rrrrrrrr}
\toprule
 &        CD-UCT &  RL-BIC &  Greedy Search & Random Search & Random Sampling &     CAM &  LiNGAM &  NOTEARS \\
\midrule
Reward $\uparrow$ & 303.466\tiny{$\pm2.532$} &     $\infty$ &        275.519 &  81.903\tiny{$\pm1.281$} &    14.270\tiny{$\pm2.490$} & 222.342 & 190.434 &   87.081 \\
   TPR $\uparrow$ &   0.852\tiny{$\pm0.009$} &     $\infty$ &          0.841 &   0.100\tiny{$\pm0.006$} &     0.046\tiny{$\pm0.006$} &   0.611 &   0.381 &    0.124 \\
   FDR $\downarrow$ &   0.148\tiny{$\pm0.009$} &     $\infty$ &          0.159 &   0.900\tiny{$\pm0.006$} &     0.954\tiny{$\pm0.006$} &   0.623 &   0.865 &    0.500 \\
   SHD $\downarrow$ &  30.880\tiny{$\pm1.704$} &     $\infty$ &         34.000 & 198.020\tiny{$\pm1.243$} &   210.220\tiny{$\pm1.353$} & 149.000 & 326.000 &  107.000 \\
\bottomrule
\end{tabular}

%% file: tex/6_conclusion.tex
\section{CONCLUSION}

\textbf{Summary.} In this work, we have proposed Causal Discovery Upper Confidence Bound for Trees (CD-UCT), a practical yet rigorous method for causal structure discovery based on an incremental construction process through Reinforcement Learning. The latter is an attractive paradigm due to its flexibility regarding the types of random variables, data generation models, and score functions it can accommodate. We have also derived an algorithm for efficiently determining cycle-inducing edges as construction progresses, which substantially improves running times. We have demonstrated significantly better performance than RL-BIC and Greedy Search on two real-world tasks, while showing that RL-BIC often performs worse than a Random Search in DAG space. Furthermore, we have examined the impact of graph density and dataset size on the relative performance of the methods and shown that, unlike RL-BIC, our method can operate on graphs with $d=50$ nodes.

\textbf{Implications.} Our results highlight that model-based RL can substantially outperform model-free RL techniques for causal discovery, echoing findings in other applications for which they have been compared. Additionally, we also showed that the myopic horizon of Greedy Search may be suboptimal, and a deeper search can yield better causal structures in a variety of settings. Given that greedy approaches are a component in many algorithms -- e.g., they are used to search the space of equivalence classes in GES~\citep{chickering2002optimal} and the space of orderings in CAM~\citep{buehlmann2014cam}, performing a deeper search may find broader uses with different state and action space formulations. 

\textbf{Future work.} An interesting research direction is the improvement of the simulation policy at the basis of our approach. Currently, it treats each edge as equally likely when performing rollouts, which is free from bias but may result in noisy estimates of future rewards. Prioritizing edges according to statistical relationships present in the dataset or external knowledge provided by an expert may improve the accuracy of the discovered structures.

\textbf{Other applications.} The proposed RL method can be applied beyond causal discovery, since it does not rely on assumptions about the state space and reward functions, as long as they consist of and can be applied to DAGs. As such, the method is directly applicable to other scenarios where finding DAGs that optimize an objective function is relevant.  Notably, DAGs are often used to represent \textit{dependencies}, e.g., those between tasks that must be scheduled in time-shared concurrent computer systems~\citep{mao2019learning}, relationships between software packages~\citep{martin2000design}, or components in a manufacturing pipeline~\citep{borenstein2000directed}.

%% file: tex/supplementary.tex
\title{Tree Search in DAG Space with Model-based \\Reinforcement Learning for Causal Discovery (Supplementary Material)}
\maketitle

\appendix

\section{PROOF OF THEOREM~\ref{th:cyc}}~\label{sec:thproof}
\cyc*

\begin{proof}
The goals are to prove that (1) each candidate edge $e_{x,y} \in \mathcal{C}_{\tau+1}$ would cause a cycle at time $\tau + 1$ and (2) that no other cycle-inducing candidate edge exists. In other words, establishing (1) proves necessity and (2) proves sufficiency.    

\textbf{(1) Necessity.} There are two possible cases, corresponding to the components of the union.

\textbf{Case 1:} $e_{x,y} \in \mathcal{C}_\tau$. By definition, the introduction of $e_{x,y}$ causes a cycle at time $\tau$. The fact that the sets of edges in successive timesteps are strict subsets of each other, i.e., $E_\tau \subset E_{\tau + 1}$, implies that $e_{x,y}$ would cause a cycle at time $\tau + 1$.

\textbf{Case 2:} $e_{x,y} \notin \mathcal{C}_\tau$. The candidate must belong to the second component of the union, which implies that $v_x \in \mathbf{De}(v_j)$ and $v_y \in \mathbf{An}(v_i)$. There are four subcases, arising from whether $x \eqmark j$ and $y \eqmark i$. 

Let us treat the subcase $(x \neq j, y \neq i)$, noting that the others are provable in a similar manner. By definitions of the descendant and ancestor sets, we have that there exist the paths $(v_j, \ldots, v_x)$ and $(v_y, \ldots, v_i)$. Given that edge $e_{i,j}$ already exists (having been added previously), the addition of edge $e_{x,y}$ would create the path $(v_i, v_j, \ldots, v_x, v_y, \ldots, v_i)$, which is a cycle. Hence, $e_{x,y}$ would cause a cycle at time $\tau + 1$.

\textbf{(2) Sufficiency.} Next, the goal is to prove that no other cycle-inducing candidate edge exists. We do this by contradiction: assume that there is an edge $e_{x,y} \notin \mathcal{C}_{\tau+1}$ such that its addition causes a cycle.

By the definition of $\mathcal{C}_{\tau+1}$ and De Morgan's laws for sets, we have that $e_{x,y} \notin \mathcal{C}_t \land e_{x,y} \notin \Phi_{i,j}$. The first clause implies that the candidate edge was not cycle-inducing before the addition of $e_{i,j}$ in the last step; therefore, any such cycle must contain the edge $e_{i,j}$ in addition to $e_{x,y}$. The second clause implies that $v_x \notin \mathbf{De}(v_j)$ and $v_y \notin \mathbf{An}(v_i)$, which means that there are no paths of the form $(v_j,\ldots,v_x)$ and $(v_y,\ldots,v_i)$. Let us use $v_j\nrightarrow v_x$ and $v_y \nrightarrow v_i$ to denote these unreachability constraints. Given the requirement for the cycle to contain both $e_{i,j}$ and $e_{x,y}$, the possible types of cycles are:
\begin{enumerate} 
    \item $(v_i, v_j\nrightarrow v_x, v_y \nrightarrow v_i)$, 
    \item $(v_x,v_y\nrightarrow v_i, v_j\nrightarrow v_x)$,
    \item $(v_{i=x}, v_y \nrightarrow v_{i=x})$, 
    \item $(v_{x=i}, v_j \nrightarrow v_{x=i})$, 
    \item $(v_i, v_{j=y} \nrightarrow v_i)$,
    \item $(v_x, v_{y=j} \nrightarrow v_x)$.
\end{enumerate}
This list is exhaustive because, by definition, $v_x \notin \mathbf{De}(v_j) \implies x \neq j$ and $v_y \notin \mathbf{An}(v_i) \implies y \neq i$. These paths cannot be cycles as a consequence of the unreachability constraints defined above. Therefore, the edge $e_{x,y} \notin \mathcal{C}_{\tau+1}$ does not cause a cycle, which contradicts our initial assumption.
\end{proof}

\section{MONTE CARLO TREE SEARCH AND RELATIONSHIP TO GREEDY SEARCH AND RL-BIC}~\label{sec:mcts_details}

\subsection{Search Problems and Shortcomings of Greedy Search}~\label{subsec:greedy}

Search has been one of the most widely utilized approaches for building intelligent agents since the dawn of AI~\citep[Chapters~3-4]{russell2010artificial}. Search methods construct a \textit{tree} in which the nodes are states in the MDP. Children nodes correspond to the states obtained by applying a particular action to the state at the parent node, while leaf nodes correspond to terminal states, from which no further actions can be taken. The root of the search tree is the current state. The way in which this tree is expanded and navigated is dictated by the particulars of the search algorithm.

Let us first discuss \textit{Greedy Search}. It creates, at each step, a search tree of depth equal to 1 rooted at the current state, evaluates the objective function for each of the child nodes, and picks the action corresponding to the best child node as the next action. The search is repeated with the child node as the root until a terminal state is reached. The main drawback of this method is that it may take short-sighted decisions, i.e., actions that are the best locally do not translate to a globally optimal solution when applied jointly. Despite its very shallow search horizon, it is commonly used in practice to good effect in a variety of problems~\citep{cormen2022introduction}, Greedy Equivalence Search~\citep{chickering2002optimal} being a relevant example.

To see why Greedy Search may be suboptimal, consider the following example in a causal discovery context: two sequences of candidate edges $(e_1,e_2)$ and $(e_3, e_4)$ decrease the BIC score by 1 and 2 respectively when applied jointly. However, if the decreases for the first edges $e_1$ and $e_3$ are 0.5 and 0.1 respectively, a greedy algorithm would not choose the second trajectory due to its myopic horizon.

\subsection{Monte Carlo Tree Search}~\label{subsec:mcts}

If greedy methods have this shortcoming, why not simply go deeper? In many applications, the branching factor and depth of the search tree make it impossible to explore all paths in the MDP. To break the curse of dimensionality, one option is to use Monte Carlo rollouts: to estimate the goodness of an action, run simulations from a tree node until reaching a terminal state~\citep{abramson_expected-outcome_1990,tesauro_online_1997}.

Monte Carlo Tree Search (MCTS) is an algorithm based on this principle. It is a model-based planning technique that addresses the inability to explore all paths in large MDPs by constructing a policy from the current state~\citep[Chapter~8.11]{sutton2018reinforcement}. Furthermore, it relies on the idea that the returns obtained by this sampling are informative for deciding the next action at the root of the search tree. We review its basic concepts below, discussing some of the notation used in the pseudocode of Algorithm~\ref{alg:cd-uct}. We also refer the interested reader to~\citep{browne_survey_2012} for a broader review of MCTS. 

In MCTS, each node in the search tree stores several statistics such as the sum of returns and the node visit count in addition to the state. For deciding each action, the search task is given a computational budget expressed in terms of node expansions or wall clock time. Specifically, we opt for the former solution and use a budget of simulations per action equal to $b_{\text{sims}} * d$ (hence linearly proportional to the number of nodes in the graph).

At each step, the algorithm keeps executing the steps below until the search budget is exhausted:

\begin{enumerate}
    \item \textbf{Selection}: The tree is traversed iteratively from the root node $n_\text{root}$ until an expandable node (i.e., a node containing a non-terminal state with yet-unexplored actions) is reached. 
    \item \textbf{Expansion}: From the expandable node, a new node is constructed and added to the search tree, with the expandable node as the parent and the child corresponding to a valid action from its associated state. This newly added node is referred to as the \textit{border node} $n_\text{border}$. The mechanism for selection and expansion is called \textit{tree policy} (denoted \textsc{TreePolicy}), and it is typically based on the node statistics.
    \item \textbf{Simulation}: Trajectories in the MDP are sampled from the border node until a terminal state is reached and the return is recorded. The \textit{default policy} or \textit{simulation policy} (denoted \textsc{SimPolicy}) dictates the probability of each action. The simplest version, which we opt for, is to use uniform random sampling of valid actions. We note that the intermediate states encountered when performing this sampling are not added to the search tree.
    \item \textbf{Backpropagation}: The return is backpropagated from the expanded node upwards to the root of the search tree, and the statistics of each node that was selected by the tree policy are updated.
\end{enumerate}

Once the budget is exhausted, the search step is completed and the action corresponding to the child node with the highest reward (denoted \textsc{MaxChild}) is chosen as the next root node. This process is repeated until a terminal state is encountered.

The tree policy used by the algorithm trades off exploration and exploitation in order to balance actions that are already known to lead to high returns against yet-unexplored paths in the MDP for which the returns are still to be estimated. The exploration-exploitation trade-off has been widely studied in the multi-armed bandit setting, which may be thought of a single-state MDP. A representative method is the Upper Confidence Bound (UCB) algorithm~\citep{auer_finite-time_2002}, which computes confidence intervals for each action and chooses, at each step, the action with the largest upper bound on the reward, embodying the principle of optimism in the face of uncertainty. 

Upper Confidence Bounds for Trees (UCT)~\citep{kocsis_bandit_2006} is a variant of MCTS that applies the principles behind UCB to the tree search setting. Namely, the selection decision at each node is framed as an independent multi-armed bandit problem. At decision time, the \textit{tree policy} of the algorithm selects the child node corresponding to action $a$ that maximizes
\begin{align}
UCT(s,a) = \bar{r_a} + 2  \epsilon_{\text{\tiny{UCT}}}   \sqrt{\frac{2  \ln{C(s)}}{C(s,a)}},
\end{align} where $\bar{r_a}$ is the mean reward observed when taking action $a$ in state $s$, $C(s)$ is the visit count for the parent node, $C(s,a)$ is the number of child visits, and $\epsilon_{\text{\tiny{UCT}}}$ is a constant that controls the level of exploration. 

UCT has been shown to converge to the optimal action with probability 1 as the number of samples grows to infinity~\citep{kocsis_bandit_2006}, and hence features a similar guarantee to Greedy Equivalence Search. However, we note that the number of samples in this context refers to the number of MC simulations that the algorithm is allowed to perform, and not the size $n$ of the considered dataset.

\subsection{Model-based versus Model-free: Why and How CD-UCT Outperforms RL-BIC}~\label{subsec:rlbic}

In this subsection, we provide the reader with an additional comparative analysis of the performance of CD-UCT versus RL-BIC and, more generally, of \textit{model-based} versus \textit{model-free} approaches.
Let us first review the distinction between model-based and model-free methods. The former category assumes knowledge of the MDP, while the latter requires only samples of agent-environment interactions. To be specific, in this context, a \textit{model} $\mathcal{M}=(P, R)$ refers to knowing, or having some estimate of, the transition and reward functions $P, R$. Given our MDP formulation of DAG construction for causal discovery (Section~\ref{subsec:mdp}), in this context we have access to the ``true'' $P$ and $R$ via their mathematical descriptions.

Model-based methods are able to exploit knowledge of $\mathcal{M}$ directly, whereas model-free methods do not. Intuitively, they can determine precisely what the subsequent states and rewards will be for a sequence of actions, without needing to go through a trial-and-error process in the environment, as model-free methods do. Therefore, given an accurate model, model-based algorithms are able to arrive at substantially better policies given the same amount of environment interaction. 

Let us give some examples of problems for which the two have been compared and results reflect these characteristics. Most relevantly,~\citet{darvariu2023planning} considered the problem of constructing undirected graphs, finding that a model-based RL method that extends MCTS greatly improves performance and scalability over a variant of the DQN model-free algorithm.~\citet{guo_deep_2014} showed that UCT outperforms DQN on a range of Atari game environments. The results of~\citet{anthony_thinking_2017} highlight the fact that UCT greatly outperforms a REINFORCE model-free agent for the connectionist game of Hex.

Returning to the causal discovery problem and the comparative performance of CD-UCT and RL-BIC, as discussed in the paragraph ``Determining budgets'' of Section~\ref{sec:experiments}, our experiments are set up such that the budget of score function evaluations awarded to CD-UCT and RL-BIC is the same. The superiority of CD-UCT is manifested as follows: CD-UCT yields substantially better performance when given the same number of score function evaluations (as shown in Table~\ref{tab:joint-results}). Furthermore, Figure~\ref{fig:budget} shows that CD-UCT can match RL-BIC performance with two orders of magnitude fewer score function evaluations, which translates to a DAG of the same quality being found in minutes instead of hours.

\section{FURTHER EXPERIMENT DETAILS}\label{sec:details}

\paragraph{Code and data.} In a future version, our implementation, data, and instructions will be publicly released in order to ensure full reproducibility of all the reported results. For both \textit{Sachs} and \textit{SynTReN}, we use the files released by~\citet{lachapelle2020gradient} at \href{https://github.com/kurowasan/GraN-DAG/tree/master/data}{this URL} under the MIT license. We leverage the synthetic data generator implementations used in~\citet{lachapelle2020gradient} for GP regression (MIT license) and~\citet{zhu2020causal} (Apache license) for quadratic regression, respectively. For discrete RV data, we use the Bayesian Network Repository~\citep{bnrepository}. 

\paragraph{Implementation details.} We build our incremental DAG construction environment and the agents including CD-UCT in pure Python. The current implementation is a prototype and substantial improvements in speed (easily 2-3 orders of magnitude) can be obtained through implementation in a lower-level programming language, especially as graph size grows. We leverage the score function, metrics, and pruning implementations released by~\citet{zhu2020causal} at \href{https://github.com/huawei-noah/trustworthyAI/tree/master/research/Causal%20Discovery%20with%20RL}{this URL} under the Apache license. For LiNGAM and NOTEARS, we use the open-source implementations provided by their respective authors. For CAM, we use the ``bridge'' in the Causal Discovery Toolbox library~\citep{kalainathan2020causal}. 

\paragraph{Parameters.} For CD-UCT, we vary the exploration parameter $\epsilon_{\text{\tiny{UCT}}} \in \{0.025, 0.05, 0.075, 0.1 \}$, while for RL-BIC we vary the model input dimension $\in \{64,128\}$ and learning rate $\in \{0.001, 0.0001\}$ -- this set includes the default parameters reported in the paper. For the other methods, the default hyperparameters are used. For CD-UCT, we use the full search horizon on \textit{Sachs} and the $d=10$ synthetic continuous RV graphs, and a reduced horizon of $h=16$ for \textit{SynTReN}, $h=8$ for the $d=50$ synthetic graph, $h=4$ for the discrete RV graphs respectively. These values were selected by grid search for reward maximization as described below. As the breadth and the depth of search tree grow, a limited horizon is required due to greatly increased search spaces, which lead to substantially noisier estimates of future rewards using uniform random sampling. For the synthetic graph experiments with varying number of datapoints $n$, the full set of considered values is $n \in \{10,25,50,75,100,175,250,375,500,750,1000,1750,2500,3750,5000\}$.

\paragraph{Hyperparameter selection methodology.} We perform a simple grid search to determine the best values of the hyperparameters. We note that the goal of hyperparameter selection is to optimize the reward (i.e., find the best graph w.r.t. $\mathcal{F}$) and not the final evaluation metric (e.g., SHD) as this would mean validating on the ground truth graph, which is methodologically incorrect. Furthermore, the initializations of the algorithms differ between the parameter tuning and evaluation runs, so that we do not simply memorize the best results.

\paragraph{Baselines.} We now provide a high-level overview of the baseline methods used in the evaluation.

\begin{itemize}
    \item \textbf{Causal Additive Models (CAM).} CAM~\citep{buehlmann2014cam} is an order-based method that assumes the Additive Noise Model. In the first stage, an ordering of the causal variables is determined using maximum likelihood estimation. In the second stage, sparse regression is used to select edges that are consistent with the ordering. 
    \item \textbf{Linear Non-Gaussian Acyclic Model (LiNGAM).} LiNGAM~\citep{shimizu2006linear} assumes a linear model with noise generated according to a non-Gaussian distribution. The technique is based on Independent Component Analysis (ICA), which is used to obtain a ``mixing matrix". After appropriate permutation and normalization, it is used to estimate pairwise ``connection strengths'' that determine edge existence.
    \item \textbf{Non-combinatorial Optimization via Trace Exponential and Augmented lagRangian for Structure learning (NOTEARS).} NOTEARS~\citep{zheng2018dags}, also discussed in the main text, proposes to replace the discrete acyclicity constraint with a continuous one. In this way, the causal identification problem is cast as a mathematical program, for which continuous optimization techniques can be used. It assumes a linear structural equation model and learns a matrix of pairwise weights that is thresholded to determine directional relationships. 
    \item \textbf{Random Search.} This method samples valid actions uniformly at random starting with an empty DAG until a terminal state is encountered and the resulting DAG is scored. The best-scoring DAG found across all performed simulations is output as the result. Since it is given a budget of simulations comparable to CD-UCT and RL-BIC, it gauges the effectiveness with which the search space is navigated.
    \item \textbf{Random Sampling.} This strategy corresponds to choosing a graph uniformly at random out of the state space. It is not intended to be competitive and only meant as a comparison point given the metrics are on different scales.
    \item \textbf{GOBNILP.} GOBNILP~\citep{cussens2012bayesian} is an exact method for learning a Bayesian network structure. It is based on a branch-and-cut approach that features an Integer Programming formulation together with cutting planes that are specifically designed for this problem.
\end{itemize}

\paragraph{Computational infrastructure.} Experiments were conducted exclusively on CPUs on an in-house High Performance Computing (HPC) cluster. On this infrastructure, the experiments for \textit{Sachs} took approximately 43 days of single-core CPU time, while the experiments for \textit{SynTReN} took approximately 1660 days ($\approx$ 4.5 years) of single-core CPU time. These figures include hyperparameter tuning. The majority of the computational time for \textit{SynTReN} was used by RL-BIC due to its inefficiency (see the ``Runtime analysis'' paragraph in Section~\ref{sec:addresults} of this Supplementary Material). The experiments for the graphs with discrete RVs took approximately 33 days of single-core CPU time.

\section{ADDITIONAL RESULTS}\label{sec:addresults}

\textbf{Runtime analysis.} In Table~\ref{tab:timings_diffvar}, we show the wall clock runtimes associated to the results presented in Table~\ref{tab:joint-results} in the main text. CD-UCT yields similar runtimes to RL-BIC on the \textit{Sachs} dataset, but it is more than an order of magnitude faster on \textit{SynTReN} despite not being implemented in a low-level programming language. This is a consequence of the targeted exploration of the search space, which leverages the decomposability of BIC to avoid regressions for already-seen parent sets of a given variable. Random Search displays the same inefficiency in terms of exploration and requires $5\times$ more time despite performing $10\times$ less simulations than CD-UCT on  the \textit{SynTReN} dataset. The methods that do not perform the same number of score function evaluations or are based on a different paradigm altogether are characterized by a lower wall clock time, partially due to implementation concerns. Finally, we note that the runtimes of the other methods that construct graphs incrementally (Greedy Search, Random Search, Random Sampling) also benefit from the incremental algorithm for determining cycle-inducing edges, as they share the same environment for graph construction. 

\textbf{Different score function.} Results with the $\mathcal{F}_{\mathrm{EV}}$ score function are shown in Table~\ref{tab:joint-results-samevar}. Corresponding runtimes are shown in Table~\ref{tab:samevart}. CD-UCT is still the best method in terms of rewards for both datasets. Interestingly, in the case of the \textit{Sachs} dataset, the best score does not necessarily lead to the best values for the metrics (TPR/FDR/SHD), with CD-UCT, RL-BIC, and Random Search all performing similarly in this regard.

\begin{table*}[t]
\caption{Approximate wall clock running time in hours, minutes, and seconds with the $\mathcal{F}_{\mathrm{HV}}$ score function.}
\label{tab:timings_diffvar}
\begin{center}
\begin{small}
\resizebox{0.87\textwidth}{!}{
\input{results/timings_final_diffvar}
}
\end{small}
\end{center}
\end{table*}

\begin{table*}[t]
\caption{Approximate wall clock running time in hours, minutes, and seconds with the $\mathcal{F}_{\mathrm{EV}}$ score function.}
\label{tab:samevart}
\begin{center}
\begin{small}
\resizebox{0.95\textwidth}{!}{
\input{results/timings_final_samevar}
}
\end{small}
\end{center}
\end{table*}

\begin{table*}[t]
\caption{Results obtained by the methods in the construction phase (top) and after undergoing pruning (bottom)  with the $\mathcal{F}_{\mathrm{EV}}$ score function.}
\label{tab:joint-results-samevar}
\begin{center}
\begin{small}
\resizebox{1\textwidth}{!}{
\input{results/samevar_joint_final}
}
\end{small}
\end{center}
\end{table*}

\newpage

\definecolor{myred}{RGB}{215, 77, 46}
\definecolor{myblue}{RGB}{126, 166, 224}

\begin{figure*}[p]
    \centering
  \includegraphics[width=0.8\textwidth,trim={0 0 0 0},clip=true]{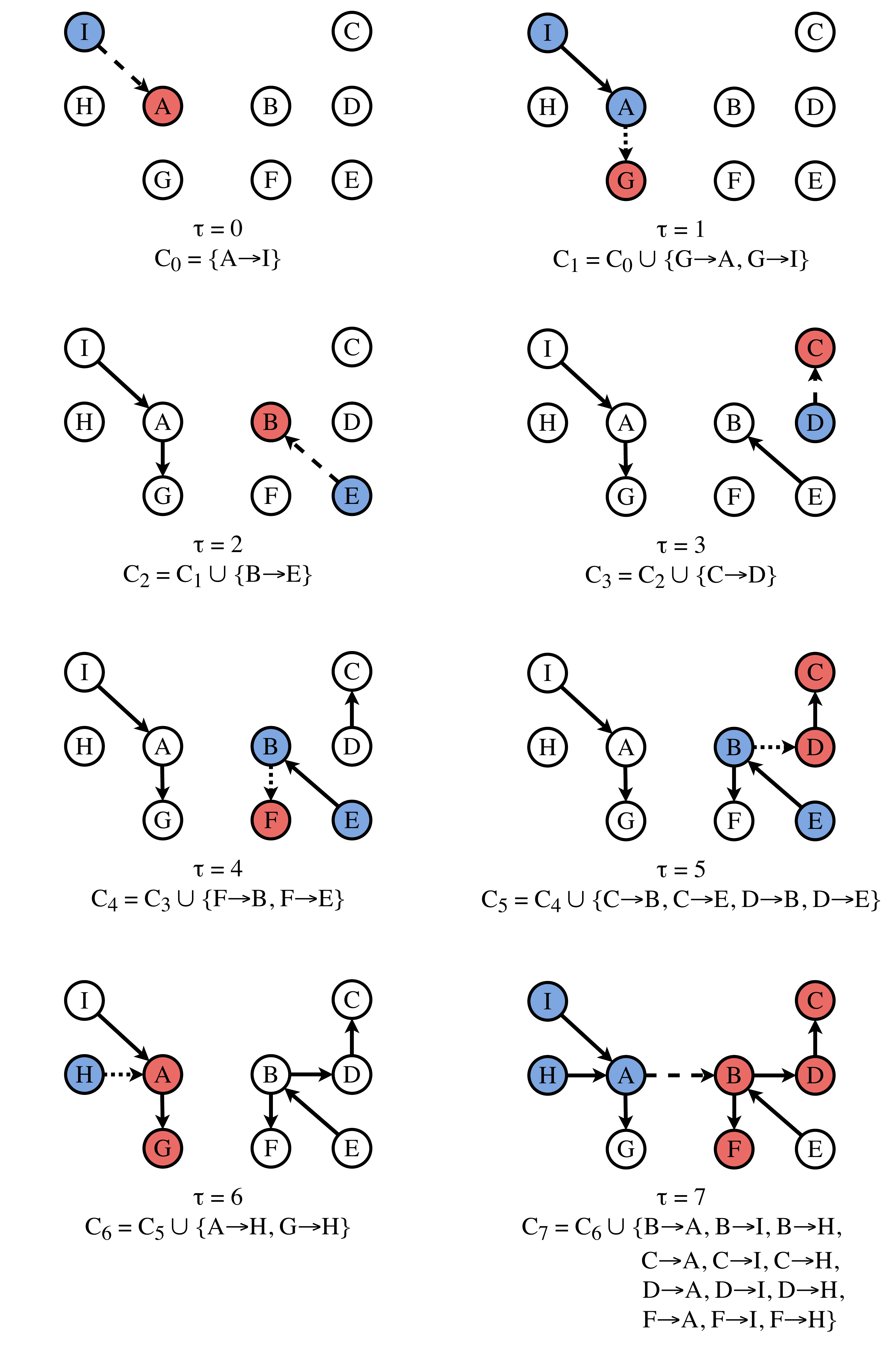}
     \caption{A full illustration of the proposed algorithm for tracking cycle-inducing edges. At each timestep $\tau$, an edge (shown with a dashed line) is introduced, and the candidate edges that would connect a \textbf{\textcolor{myred}{descendant}} of the endpoint to an \textbf{\textcolor{myblue}{ancestor}} of the starting point are added to the set $\mathcal{C}_\tau$. This eliminates the need to explicitly check for cycles. We also note that this algorithm simply determines the \textit{invalid} edges, with the choice of which edge to add being left to the higher-level causal discovery method.} 
  \label{fig:incremental_full_run}
\end{figure*}

\vfill

%% file: results/timings_final_diffvar.tex
\begin{tabular}{c|rrrrrrrr}
\toprule
 &   CD-UCT &    RL-BIC & Greedy Search & Random Search & Random Sampling &      CAM &   LiNGAM &  NOTEARS \\
\midrule
   \textit{Sachs} &  1:53:24 &   1:43:33 &       0:01:48 &       3:46:56 &           0:00:04 &  0:00:39 &  0:00:04 &  0:00:05 \\
 \textit{SynTReN} &  2:10:37 &  35:32:20 &       0:03:11 &      10:35:48 &           0:00:07 &  0:01:38 &  0:00:22 &  0:01:58 \\
\bottomrule
\end{tabular}

%% file: results/timings_final_samevar.tex
\begin{tabular}{c|rrrrrrrr}
\toprule
 &   CD-UCT &    RL-BIC & Greedy Search & Random Search & Random Sampling &      CAM &   LiNGAM &  NOTEARS \\
\midrule
   \textit{Sachs} &  1:49:39 &   1:30:12 &       0:01:20 &       2:35:02 &         0:00:02 &  0:00:27 &  0:00:03 &  0:00:03 \\
 \textit{SynTReN} &  2:01:43 &  35:50:42 &       0:03:07 &       9:33:47 &         0:00:06 &  0:01:39 &  0:00:22 &  0:01:59 \\
\bottomrule
\end{tabular}

%% file: results/samevar_joint_final.tex
\begin{tabular}{ccc|rrrrrrrr}
\toprule
     Phase &  &   &        CD-UCT &         RL-BIC &  Greedy Search & Random Search & Random Sampling &      CAM &   LiNGAM &  NOTEARS \\
\midrule
 \textbf{Construct} &    \textit{Sachs} &  Reward $\uparrow$ &  -0.711\tiny{$\pm0.002$} &   -0.726\tiny{$\pm0.035$} &         -0.756 &  -0.748\tiny{$\pm0.003$} &    -1.409\tiny{$\pm0.018$} & --- & --- & --- \\
 &    &     TPR $\uparrow$ &   0.524\tiny{$\pm0.015$} &    0.565\tiny{$\pm0.015$} &          0.471 &   0.558\tiny{$\pm0.018$} &     0.421\tiny{$\pm0.030$} & --- & --- & --- \\
&     &     FDR $\downarrow$ &   0.818\tiny{$\pm0.005$} &    0.814\tiny{$\pm0.004$} &          0.837 &   0.806\tiny{$\pm0.006$} &     0.854\tiny{$\pm0.010$} & --- & --- & --- \\
&     &     SHD $\downarrow$ &  42.070\tiny{$\pm0.438$} &   43.250\tiny{$\pm0.689$} &         41.000 &  41.220\tiny{$\pm0.438$} &    43.700\tiny{$\pm0.587$} & --- & --- & --- \\
\midrule
&  \textit{SynTReN} &  Reward $\uparrow$ &  -0.296\tiny{$\pm0.011$} &   -0.835\tiny{$\pm0.101$} &         -0.361 &  -0.506\tiny{$\pm0.017$} &    -1.302\tiny{$\pm0.038$} & --- & --- & --- \\
 &   &     TPR $\uparrow$ &   0.438\tiny{$\pm0.021$} &    0.318\tiny{$\pm0.032$} &          0.332 &   0.317\tiny{$\pm0.015$} &     0.256\tiny{$\pm0.014$} & --- & --- & --- \\
 &   &     FDR $\downarrow$ &   0.894\tiny{$\pm0.005$} &    0.926\tiny{$\pm0.006$} &          0.917 &   0.923\tiny{$\pm0.004$} &     0.938\tiny{$\pm0.004$} & --- & --- & --- \\
\rule{0pt}{2ex}   
 &   &     SHD $\downarrow$ &  93.565\tiny{$\pm0.938$} &  102.075\tiny{$\pm5.550$} &         96.300 &  99.805\tiny{$\pm0.833$} &   102.155\tiny{$\pm0.899$} & --- & --- & --- \\
\midrule[0.175em]
\rule{0pt}{2ex}   
     \textbf{Prune} &    \textit{Sachs} &     TPR $\uparrow$ &   0.334\tiny{$\pm0.013$} &    0.334\tiny{$\pm0.011$} &          0.235 &   0.339\tiny{$\pm0.016$} &     0.211\tiny{$\pm0.021$} &    0.353 &    0.235 &    0.118 \\
      &     &     FDR $\downarrow$ &   0.388\tiny{$\pm0.022$} &    0.331\tiny{$\pm0.018$} &          0.600 &   0.369\tiny{$\pm0.027$} &     0.577\tiny{$\pm0.039$} &    0.400 &    0.556 &    0.500 \\
      &     &     SHD $\downarrow$ &  12.150\tiny{$\pm0.217$} &   11.730\tiny{$\pm0.199$} &         14.000 &  12.040\tiny{$\pm0.257$} &    14.060\tiny{$\pm0.352$} &   12.000 &   14.000 &   15.000 \\
\midrule      
      &  \textit{SynTReN} &     TPR $\uparrow$ &   0.297\tiny{$\pm0.017$} &    0.168\tiny{$\pm0.015$} &          0.234 &   0.234\tiny{$\pm0.014$} &     0.189\tiny{$\pm0.012$} &    0.329 &    0.303 &     $\times$  \\
      &   &     FDR $\downarrow$ &   0.822\tiny{$\pm0.010$} &    0.873\tiny{$\pm0.011$} &          0.854 &   0.856\tiny{$\pm0.009$} &     0.879\tiny{$\pm0.008$} &    0.794 &    0.809 &     $\times$  \\
      &   &     SHD $\downarrow$ &  44.020\tiny{$\pm1.119$} &   43.500\tiny{$\pm1.167$} &         45.800 &  46.890\tiny{$\pm1.120$} &    47.715\tiny{$\pm0.946$} &   41.300 &   41.600 &    $\times$  \\
\bottomrule
\end{tabular}